\documentclass[lettersize,journal]{IEEEtran}
\usepackage{graphicx} 
\usepackage{amsmath,amsfonts}
\usepackage{algorithmic}
\usepackage{algorithm}
\usepackage{array}
\usepackage[caption=false,font=normalsize,labelfont=sf,textfont=sf]{subfig}
\usepackage{textcomp}
\usepackage{stfloats}
\usepackage{url}
\usepackage{verbatim}
\usepackage{cite}
\usepackage{longtable,tabularx}
\usepackage{siunitx}
\usepackage{wrapfig}
\usepackage{xcolor}
\usepackage[export]{adjustbox}
\usepackage[utf8]{inputenc}
\usepackage[T1]{fontenc}
\begin{document}

\title{Design and Characterization of a Unified Continuum Robot Actuation Platform}

\author{Yunti Xu, Aedan Mangan, Miheer Potdar, Alex Lu, and Tania K. Morimoto,~\IEEEmembership{Senior Member,~IEEE}
\thanks{Yunti Xu, Aedan Mangan, Miheer Potdar, Alex Lu, and Tania K. Morimoto are with the Department of Mechanical and Aerospace Engineering, University of California, San Diego, CA, 92093, USA. {\tt\small ytxu@ucsd.edu}, {\tt\small amangan.ucsd.edu}, {\tt\small mpotdar@ucsd.edu},{\tt\small allu@ucsd.edu}, {\tt\small tamorimoto@ucsd.edu}}

}
\markboth{Journal of \LaTeX\ Class Files,~Vol.~14, No.~8, August~2021}%
{Shell \MakeLowercase{\textit{et al.}}: A Sample Article Using IEEEtran.cls for IEEE Journals}

\maketitle

\begin{abstract}
Continuum robots enable dexterous access to confined spaces in applications such as minimally invasive surgery and industrial inspection. However, their actuation systems are often custom-designed for specific robot architectures and tasks, limiting reuse across publications, institutions, and applications. This paper presents a unified actuation platform that decouples actuation hardware and software from continuum robot design. The platform provides modular actuation across continuum robot classes, scales to a large number of degrees of freedom (DOFs), and integrates with conventional serial manipulators for hybrid architectures. We validate the platform on an open-sourced surgical robotic system and experimentally characterize the actuation modules, achieving sub-millimeter translational accuracy and less than $5^\circ$ rotational error over ranges of $30$~mm and $180^\circ$, respectively. We further demonstrate the platform in two surgical use cases: teleoperation of a continuum camera in a constrained environment and bimanual manipulation using $2.53$~mm concentric-tube-robot-based forceps alongside a conventional wristed instrument. All hardware designs and ROS-based software are open-sourced to facilitate reuse and further research in continuum and surgical robotics.
\end{abstract}





\vspace{-8pt}
\section{Introduction} \label{sec_intro}

\begin{figure}[t!]
\centerline{\includegraphics[width=1.0\linewidth]{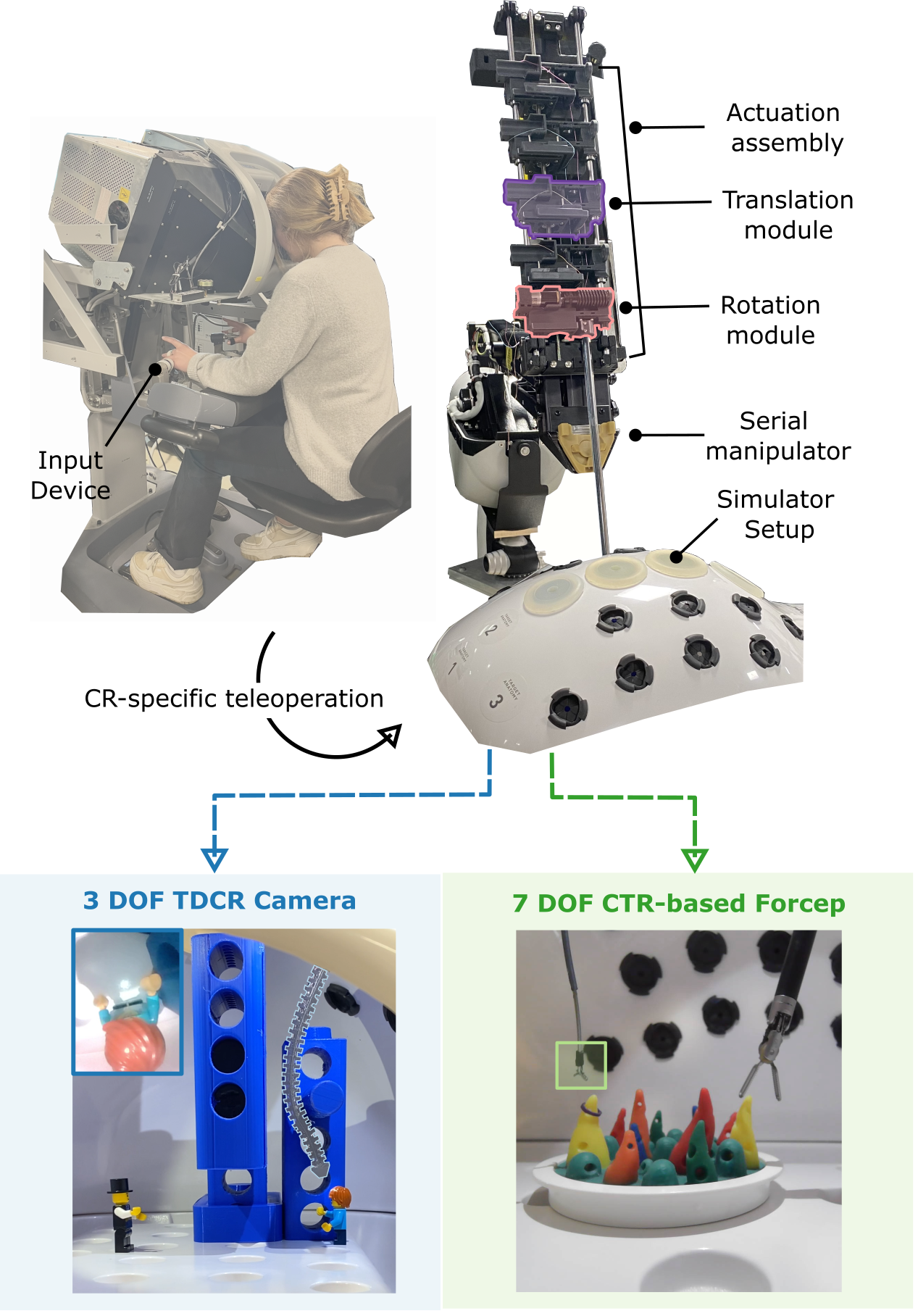}}
\vspace{-8pt}
\caption{Overview of the unified continuum robot actuation platform on a surgical robotic system, showcasing two teleoperation examples of different potential applications with various robot configurations. For better visualization, we apply false coloring to the TDCR camera.}
\label{fig_cover}
\end{figure}


\IEEEPARstart{C}{ontinuum} robots, with their continuously bending structures and small size, are well suited for navigating constrained and cluttered environments~\cite{JBK2015}, making them promising platforms for minimally invasive surgery and industrial inspection~\cite{JBK2015,cr-review}. However, researchers typically develop custom, proprietary actuation hardware and software for specific robot designs and tasks, limiting reuse across publications, institutions, and applications. For example, a study of concentric tube robots found that $61.1\%$ of prototypes appeared in only a single publication, with similar trends observed across other continuum robot classes~\cite{grassmann2020ctcr}. The need for a generalized actuation platform is further highlighted by the diversity of existing continuum robot classes and configurations~\cite{JBK2015}, as well as hybrid systems that integrate continuum robots with conventional rigid manipulators~\cite{ma2022-IROS,Zhang2020-serial-endoscope,Zhang2025-tdcr-on-UR-TMRB,Qi2025-tdcr-on-serial-neurosurgery,Dwyer2017-ctr-on-serial-fetoscope,Francis2017-CTR-dVRK-RAM,ma2023design,shentu2026samplingbasedfollowtheleadermotionplanning}. In this work, we present a unified, robot design-agnostic actuation platform for continuum robots that can easily integrate with other robotic systems.






\subsection{Design of Continuum Robot Actuation Systems}\label{subsec_continuum_robot_actuation_design}
Although the working principles of continuum robots can vary, their actuation commonly relies on a combination of three main modes: linear translation, axial rotation, and tendon-induced motion.
For tendon-driven continuum robots (TDCRs), tendon displacement can be generated directly using lead-screw mechanisms~\cite{Xu2008TDCR,Amanov2021,Sarma2022COAST} or through capstan drives~\cite{mangan2025serially,lin2025hcr,Zhang2020-serial-endoscope,ma2022-IROS}. Concentric tube robots (CTRs)~\cite{SearsDupont2006,Webster2006} are actuated through the relative translation and axial rotation of nested, pre-curved Nitinol (NiTi) tubes, typically with diameters on the order of $1$--$2$~mm. These motions can likewise be realized using different mechanisms, including gear-based~\cite{morimoto2017,Farooq2017CTR}, belt-based~\cite{Hendrick2014CTR}, and screw-based~\cite{Esakkiappan2019} mechanisms for linear translation, and gear-based~\cite{morimoto2017,Hendrick2014CTR}, and pulley-based~\cite{Esakkiappan2019,Yu2016CTR} mechanisms for axial rotation.

Different modes of actuation impose different actuation requirements. For TDCRs, small inaccuracies in tendon displacement can produce large unintended motions at the distal tip~\cite{bajo2011kinematics, Zhang2025-tdcr-on-UR-TMRB, mangan2025serially, lin2025hcr}, requiring accurate tendon displacement and sufficient actuation force~\cite{bajo2011kinematics, Zhang2025-tdcr-on-UR-TMRB, mangan2025serially, Qi2025-tdcr-on-serial-neurosurgery, ma2023design}. For CTRs, high tube pre-curvature can generate substantial axial and torsional friction during relative translation and rotation~\cite{lin2024closing}, requiring sufficient force and torque while maintaining positioning accuracy under load~\cite{morimoto2017}. Together, these differences in actuation requirements and mechanism design make a custom actuation system developed for one continuum robot difficult to transfer directly to another. 

Grassmann et al.~\cite{JBK2024-open-continuum} addressed this challenge with a modular actuation system built around a common rotational module. Mode-specific mechanisms convert rotational input into tendon displacement for TDCRs or tube translation and axial rotation for CTRs. This work uses the same drive modules to actuate a 4-DOF CTR, a 2-DOF planar TDCR, and a 3-DOF spatial TDCR, with minimal module hardware changes between robot classes. However, the number and arrangement of drive modules vary with the robot design, resulting in substantially different actuation architectures and physical footprints. Thus, while the drive modules themselves are reusable between actuation modes, the actuation architecture remains configuration-dependent and must be re-conceived for each robot design. This dependence makes scaling the actuation system to higher-DOF configurations more challenging and can limit its mechanical integration with other robotic systems.

\subsection{Hybrid Continuum Serial Arm Robots}\label{subsec_hybrid_continuum_serial_arm_robots}
To date, researchers have combined continuum robots with serial manipulators in both surgical and general robotics applications. For surgical applications, Ma et al.~\cite{ma2022-IROS} and Zhang et al.~\cite{Zhang2020-serial-endoscope} mounted TDCRs on serial manipulators for laparoscopic visualization, while Zhang et al.~\cite{Zhang2025-tdcr-on-UR-TMRB} and Qi et al.~\cite{Qi2025-tdcr-on-serial-neurosurgery} developed TDCR-based instruments for neurosurgical and spinal applications. Dwyer et al.~\cite{Dwyer2017-ctr-on-serial-fetoscope} and Francis et al.~\cite{Francis2017-CTR-dVRK-RAM} integrated CTRs with serial manipulators for fetoscopy. Beyond surgery, Ma et al.~\cite{ma2023design} combined a TDCR with a 6-DOF movable base to investigate dexterity, manipulability, and obstacle avoidance, while Shentu et al.~\cite{shentu2026samplingbasedfollowtheleadermotionplanning} incorporated a 6-DOF movable base into motion planning for a TDCR to improve follow-the-leader (FTL) behavior. Across these applications, the continuum robots ranged from two to nine DOFs and employed robot design-specific actuation assemblies. This variation highlights the need for a unified actuation platform that scales across different DOFs and configurations and integrates with conventional serial manipulators.

\subsection{Contributions} \label{sec_contributions}
We present a unified actuation platform---including the mechanical, electrical, and software components---that decouples actuation from continuum robot design, enabling:
\begin{itemize}
    \item actuation of multiple classes of continuum robots with arbitrary DOFs,
    \item deployment across many diverse use cases, such as a flexible camera or surgical instrument,
    \item hybrid continuum–serial robot architectures.
\end{itemize}
We validate the full system on a hybrid continuum–serial robot example---using the patient-side manipulator (PSM) of the da Vinci Research Kit (dVRK) as the serial arm---through two representative teleoperated surgical use cases, demonstrating plug-and-play compatibility with an existing robotic system and the potential impact in the surgical and continuum robotics field. All hardware designs and ROS-based control software are open-sourced, enabling researchers to adopt, extend, and deploy the platform.



\section{Mechanical Design} \label{sec_mechanical_design}

\begin{figure*}[t!]
\centerline{\includegraphics[width=1.0\linewidth]{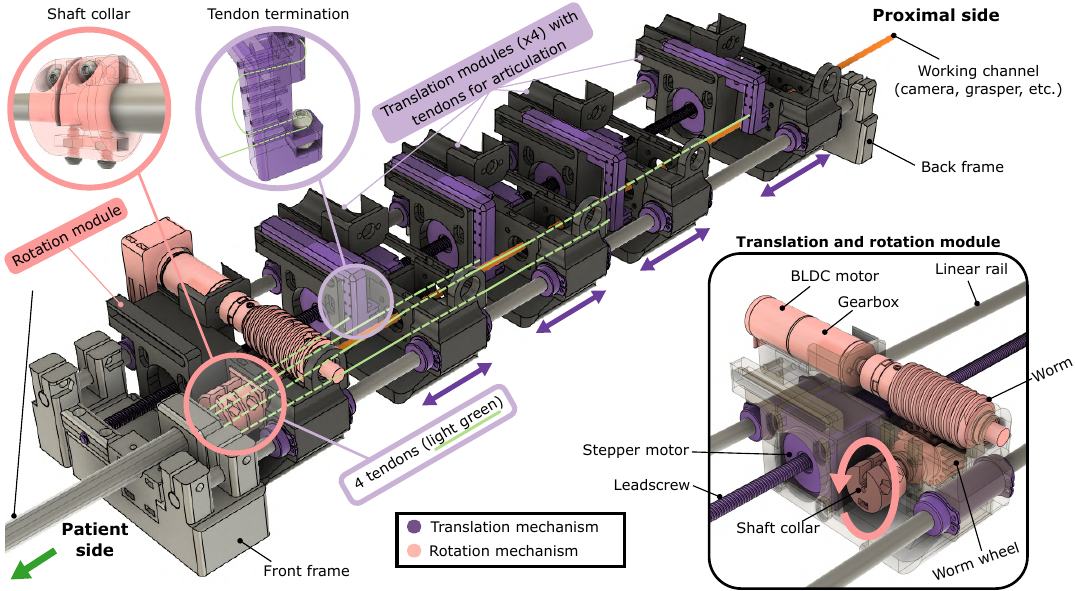}}
\vspace{-8pt}
\caption{
An example actuation assembly for a TDCR consists of five actuation modules.
The rotation mechanism (light red) is driven by a brushless DC (BLDC) motor coupled to a worm gear set aligned with the instrument axis. 
The translation mechanism (purple) is controlled by a stepper motor on a leadscrew, as the motor actuates, linear rails allow the actuation module to translate forward and backward.
The 3d-printed shaft collar can be customized to fit various instrument shaft/tube diameters based on specific robot designs.
Tendons (green, light blue, yellow, and light green) route through the assembly and terminate at individual modules, allowing for independent tendon actuation in the TDCR.
The actuation assembly design also incorporates a central working channel to accommodate cameras, forceps, and other tools. Overall, the modular architecture enables a reconfigurable actuation system that adapts to continuum robots with varying designs and numbers of actuated DOFs.
}
\label{fig_module_design}  
\end{figure*}

In this section, we first present the design requirements and an overview of the mechanical and electrical hardware components of the actuation assembly. We then detail the sub-mechanisms comprising a modular actuation unit, describe how individual units combine to form a complete assembly, and present its integration with a serial manipulator. Finally, we describe the electronics hardware.

\subsection{Design Requirements and Overview}

To accommodate different continuum robot classes and numbers of DOFs, the actuation assembly must provide modular and scalable actuation, supporting both linear translation and axial rotation while maintaining high positioning accuracy. A common motor controller must interface with all actuated DOFs to extend this modularity to the electrical hardware. In addition, to support diverse end-use applications, the assembly must independently actuate end-effectors such as graspers and forceps. Finally, for integration with hybrid continuum–serial manipulators, the assembly must be lightweight to avoid overloading the serial manipulator's actuators, maintain a compact and slim form factor for mounting, and enable rapid attachment/detachment.

To meet the above design requirements, we have designed the actuation assembly to consist of modular units that collectively provide multiple DOFs (Fig.~\ref{fig_module_design}), with each module capable of linear translation and axial rotation. The modules are integrated in series on a common frame using a lead screw and alignment rails. This modular architecture allows users to scale the total number of DoFs to accommodate different tool requirements. 
Complete assembly instructions, and sourcing information for off-the-shelf components are provided in Supplementary File~1, and the custom CAD files are hosted on \footnote{https://ucsdmorimotolab.github.io/unified-actuation/}. A video demonstration of the assembly process is available in Supplementary Video~2.

As a representative serial manipulator, we use the Patient-Side Manipulator (PSM) of the da Vinci Research Kit (dVRK), an open-source surgical robotic system for research. The PSM provides a remote-center-of-motion (RCM)-constrained arm with multiple DOFs, of which we use three—outer yaw, pitch, and insertion—to demonstrate integration with the proposed actuation platform.





\vspace{-8pt}
\subsection{Module Design} \label{sec_module design}
\subsubsection{Translation Mechanism} \label{sec_translation_mechanism}
The translation mechanism (purple in Fig.~\ref{fig_module_design}) performs the insertion and retraction of each module relative to the others, enabling both tendon actuation and tube translation.
Linear bearings constrain the module housing, allowing the module to translate along the linear rails.
For tendon actuation, a tendon routes back to the module and attaches via an \textit{L}-bracket (Fig.~\ref{fig_module_design}). 
The tendon wraps around a bolt which is fastened down, enabling quick attachment while being able to exert high forces without slipping.
For tube actuation, the module interfaces with the tube through the coupling described in Section~\ref{sec_rotation_mechanism}, allowing tube insertion.

\subsubsection{Rotation Mechanism} \label{sec_rotation_mechanism}
The rotation mechanism (light red in Fig.~\ref{fig_module_design}) enables continuous axial rotation of the instrument shaft or tubes about the insertion axis.
We use a worm gear mechanism to achieve rotation concentric to the working channel (orange in Fig.~\ref{fig_module_design}) while minimizing backlash.
An additional benefit of this design is that the worm gear is not back-drivable, allowing the motor to maintain position under load with reduced holding current, preventing motor overheating during extended working times (Section~\ref{sec_elec_design}). 
A brushless DC motor (Maxon EC-max 16) with a 19:1 planetary gearbox (Maxon, 16 A) provides sufficiently high torque, low mass, and unlimited rotation.
The motor drives a 10:1 worm gear pair composed of an off-the-shelf worm (McMaster-Carr 57545K634) and a 3D-printed worm wheel (Tough 1500 Resin V2, Formlabs Inc.).
The rotating shaft interfaces with a shaft collar (Fig.~\ref{fig_module_design}) that
users can adjust to meet the size requirements of their continuum robot.
An essential element of the design is the hollow lumen maintained throughout the actuation assembly while rotating the instrument shaft about the insertion axis, which allows for a working channel (orange in Fig.~\ref{fig_module_design}) for cameras, graspers, or other end effectors depending on the use case.

\subsubsection{Instrument Actuation Mechanism} \label{sect_instrument_actuation}
To actuate tendon-driven instruments, such as forceps, we mount a push-pull actuation mechanism to the most proximal actuation module.
After passing through the working channel, the instrument's tendons attach to a custom hex-nut adapter and couple to a BLDC-driven push-pull actuation mechanism dedicated to the tool. Details on the instrument actuation mechanism can be found in Supplementary Video~2. 

\subsection{Assembly and Attachment} \label{sec_assembly_and_dvrk_attachment}
\subsubsection{Actuation Assembly}
We attach both linear rails and the lead screw described in Section~\ref{sec_translation_mechanism} to a 3D-printed front frame (Bambu Labs PLA)(Fig.~\ref{fig_module_design}).
Each module aligns its linear bearings with the rails and inserts forward into position, repeating for the desired number of modules.
After installing all modules, the back frame attaches to complete the actuation assembly.
All frame components—including the rails, front frame, and back frame—have a combined mass of $312$~g, minimizing overall system weight and enabling mounting on serial-link robotic arms. Table~\ref{tab:system_masses} summarizes the mass of each subassembly. The complete actuation assembly measures $400 \times 98.6 \times 40$~mm (L × W × H), allowing it to fit on serial manipulators with a minimal footprint.

\subsubsection{Mechanical Interface to the Serial Manipulator} \label{sec_mechanical_interfaces_to_the_PSM}
As an example use case in surgery, we designed a modified sterile adapter to attach the actuation platform to the dVRK's Patient Side Manipulator (PSM). The front frame bolts directly to this adapter, which then mounts to the PSM through a quick-release mechanism that enables tool-free attachment and detachment in under one minute.
Since conventional wristed instruments weigh approximately 50~g~\cite{instrument_weight}, our design incorporates a constant-force spring (McMaster-Carr 9293K114) that offsets the additional weight introduced by the actuation assembly. Empirical testing shows that the chosen spring supports a 3-tube CTR assembly with an instrument attached without exceeding the PSM insertion force limits of the PSM. 
    
\begin{table}[t]
\centering
\caption{Masses of system configurations and subassemblies.}
\label{tab:system_masses}
\begin{tabular}{l c}
\hline
\textbf{Component / Configuration} & \textbf{Mass (g)} \\
\hline
CTR Module (single) & 320 \\
TDCR Roll Module (single) &  215 \\
TDCR Tendon Module (single) & 192\\
Rail System  &  312 \\
dVRK Attachment Clamp  & 260 \\
\hline
Total CTR System (3 modules + rail + CTR + instrument) & 1561 \\
Total TDCR System (5 modules + rail + TDCR) & 1328\\
\hline
\end{tabular}
\end{table}

\vspace{-8pt}
\subsection{Electronics Design} \label{sec_elec_design}
\begin{figure}[t!]
\centerline{\includegraphics[width=1.0\linewidth]{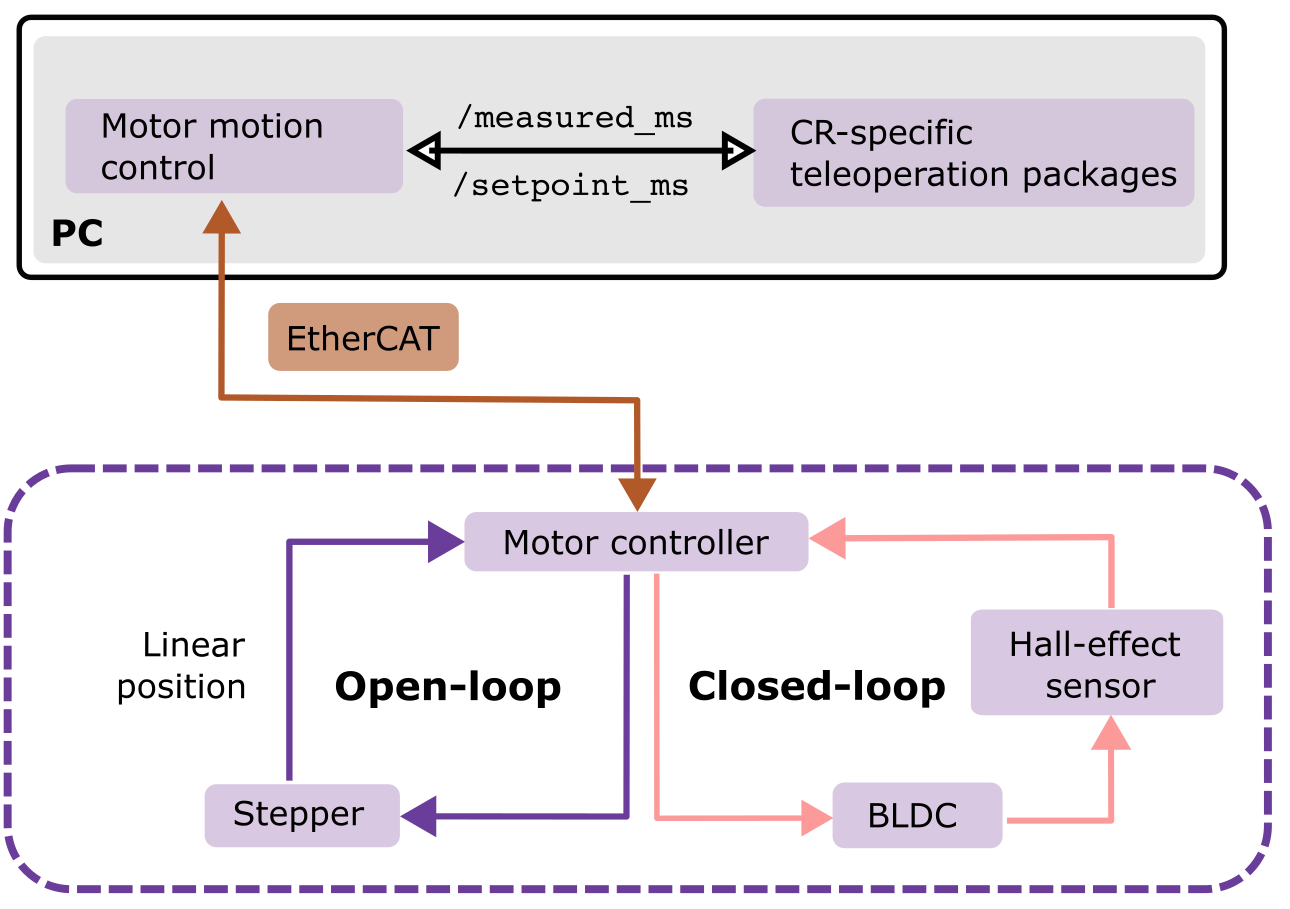}}
\vspace{-8pt}
\caption{Overview of the control pathways: a single motor controller drives either a stepper (dark purple arrows) or a BLDC (light purple arrows). The custom motor motion control directly interfaces with the motor controller, and contains a ROS2 abstraction layer which allows user to interact with the actuation/sensing components via ROS2 messages \texttt{/setpoint\_ms} and \texttt{/measured\_ms} exclusively ('\texttt{ms}' stands for motor state). These messages are then communicated to the CR-specific teleoperation packages that enable kinematic control of different continuum robotic tools}
\label{fig_motor_control}  
\end{figure}

We select a multi-axis controller board (Nanotec CM-CPB3-44-0005-1, 4-axis) that contains four independent motor controllers capable of driving either stepper or BLDC motors, thereby making the control hardware agnostic to any specific drive mechanisms used in the actuation assembly. Each controller supports multiple sensor modalities for low-level closed-loop control. Specifically, the BLDC motors employ Hall-effect sensors for position feedback, while steppers operate in open-loop (see Fig.~\ref{fig_motor_control}). The internal current control loop runs at 20~kHz.

Depending on the total number of DOFs in the actuation assembly, multiple 4-axis boards can be daisy-chained via EtherCAT, with a single board interfacing with the host PC. The controller processes data at 1~kHz. Each board operates over a voltage range of 12–57.6~V, which encompasses the operating requirements of the selected stepper and BLDC motors, both of which can operate at 12~V. The maximum current rating per controller is 3~A, which exceeds the expected peak current draw of the selected stepper and BLDC motors (1~A and 0.762~A, respectively), providing safe current margin.

\section{Software Design}
In this section, we first describe the custom motor motion control that enables application-dependent customization and seamless interfacing with any ROS-based open-sourced or commercial software stacks. Then, as part of our case study, we explain how to integrate the custom continuum robot control logic with the dVRK software stack maintained by JHU~\cite{crtk2020}. 

\subsection{Design Requirements}
To enable a unified actuation platform that decouples actuation from continuum robot design, there are two main design requirements.
First, the motion controller must support on-the-fly reconfiguration of control parameters to accommodate different actuation assemblies. Second, to facilitate integration with hybrid continuum–serial manipulator systems, the motion controller must support communication with other open-source and commercial robotic systems via ROS2.

\subsection{Motor Motion Control}
Our custom motor motion control program\footnote{https://ucsdmorimotolab.github.io/unified-actuation/} communicates directly with the controller board via HEX-encoded messages over EtherCAT. This design enables application-specific motor parameters—such as safe current limits and PID control gains—to be easily configured at program startup, while supporting continuous motor position control during runtime. The control program employs an object-oriented architecture in which each motor corresponding to a specific actuation DOF is represented by a dedicated instance that stores its parameters. This modular approach enables scalability to an arbitrary number of DOFs, accommodates multiple motor types, and allows per-DOF customization based on task-specific motion requirements. As a result, the motion controller can be readily adapted to any user-defined actuation assembly configurations. To enable flexible integration with other robotic research platforms such as the dVRK, we implemented a ROS2 abstraction layer. High-level data input/output - such as motor setpoint position and measured positions (referred to as \texttt{/setpoint\_ms} and \texttt{/measured\_ms} respectively in Fig.~\ref{fig_motor_control}) - are communicated via ROS2 topics, providing an interface between the motor motion control and higher-level control logic, such as teleoperation.

\subsection{Software Integration}
To ensure seamless integration with the existing dVRK software infrastructure, we developed custom ROS2 Python packages for CR-specific teleoperation \footnote{https://ucsdmorimotolab.github.io/unified-actuation/}. These packages (\texttt{tdcr\_teleop\_python} and \texttt{ctr\_teleop\_python}) leverage the open-sourced ROS-based Python Collaborative Robotics Toolkit~\cite{crtk2020} to access robot state information from various dVRK components and implement leader–follower teleoperation logic, inverse kinematics, and joint-to-actuator mappings specific to continuum robots.

\section{Device Evaluation}
We conducted experiments to characterize (i) the translational and rotational accuracy of the actuation modules and (ii) the impact of the attached actuation assembly on the PSM’s static range of motion as part of our use case validation. 

\subsection{Actuation Module Accuracy Characterization}
\begin{figure}[t!]
\centerline{\includegraphics[width=1.0\linewidth]{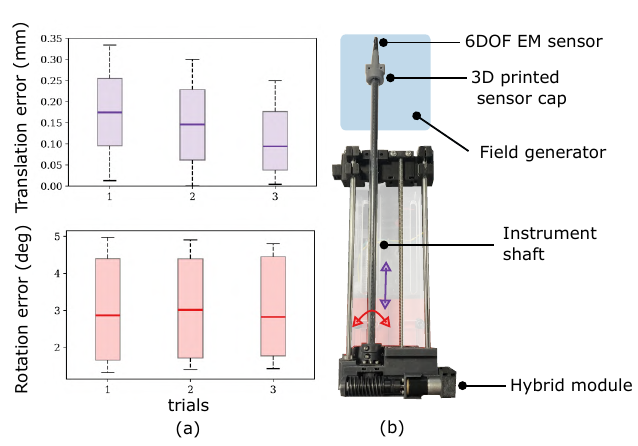}}
\vspace{-8pt}
\caption{(Left) Box plots capturing the distribution of the absolute error between the commanded position and the measured position at individual steps for each trial. (Right) Overhead view of the characterization experiment setup, the field generator is placed directly below the 6 DOF EM sensor and away from the module to minimize interference.}
\label{fig_module_accuracy} 
\end{figure}

To characterize module accuracy, we used a single module that provides both translation and rotation actuation. Fig.~\ref{fig_module_accuracy}(b) illustrates the experimental setup. An instrument shaft -- similar to that used for TDCR -- is attached to the module. At the distal end, we mounted a 3D-printed cap that secures a 6 DOF EM sensor (NDI Aurora), positioned directly above the EM field generator. 

For the translation and rotation DOFs, we commanded increments of $1$~mm and $5^\circ$ at $1$~Hz for a total of $15$~mm and $90^\circ$, respectively, and then returned to the starting position. At each step per trial, the sensor recorded the absolute error between the commanded position and the measured change in Euclidean distance or rotation angle. Fig.~\ref{fig_module_accuracy}(a) presents the translation and rotation results across three trials.

The results showed an average maximum accumulated error of $0.29 \pm 0.03$~mm for translation over the total tested distance of $30$~mm and $4.89 \pm 0.07^\circ$ for rotation over the total tested angle of $180^\circ$, with the largest errors occurring at direction reversal, where backlash is maximal. These findings demonstrated that the proposed translation mechanism achieved sub-millimeter insertion accuracy under open-loop control, while the rotation mechanism achieved angular accuracy $<5^\circ$, indicating minimal backlash in the worm gear mechanism design.

\begin{figure}[t!]
\centerline{\includegraphics[width=1.0\linewidth]{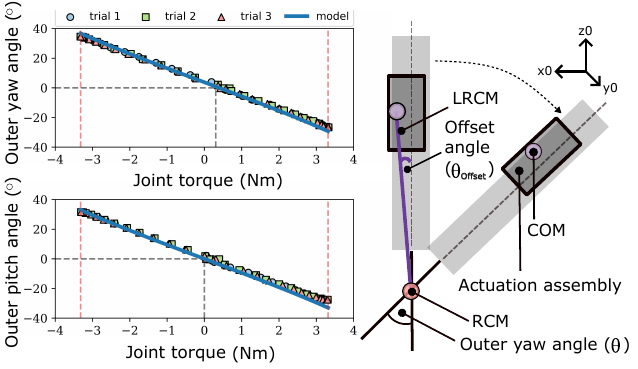}}
\vspace{-8pt}
\caption{(Left) Full range of motion span of the outer yaw and pitch joints on the PSM model prediction vs. experimental data. (Right) Schematic of the static range of motion calculation (shown only for outer yaw although outer pitch is similar). Red dashed lines indicate the torque limits set in dVRK software.}
\label{fig_rom_test} 
\end{figure}

\subsection{Range of Motion Characterization} \label{sec_exp_range_of_motion_}

Unlike for the insertion axis, the modified sterile adapter does not offer a weight-offset mechanism for the PSM's outer yaw and pitch axes, i.e. rotation about the $x0$ and $y0$ axes in Fig.~\ref{fig_rom_test} respectively, as described in Section~\ref{sec_assembly_and_dvrk_attachment}. Consequently, we characterized the impact of the actuation assembly on the PSM range of motion by determining new limits for the outer pitch and yaw joint angles. We define the following static relationship between the corresponding joint torques and joint angles: 


\begin{align}
    \theta = \sin^{-1}\left(\frac{-\tau}{mgl_{RCM}}\right) + \theta_{offset},
\end{align}

\noindent where $l_{RCM}$ denotes distance between the remote center of motion (RCM), a mechanical characteristic of the PSM, to the center of mass (COM) of the actuation assembly; $\tau$ and $\theta$ represent the joint torque and angle; $\theta_{offset}$ is an offset angle defined by the displacement of the COM from the insertion-axis centerline, which intersects the RCM; $m$ is the mass of a given actuation assembly; and $g$ is the gravitational acceleration. Using this relationship, we determined the upper and lower joint angles at which each joint reaches its torque limit.

We experimentally evaluated the outer pitch and yaw limits using the 3-tube CTR actuation assembly, which is the heaviest configuration (see Table~\ref{tab:system_masses}). We considered a configuration in which all modules were stacked close to the front frame and the entire actuation assembly was retracted along the PSM insertion axis as much as possible. Specifically, we incrementally increased and decreased the PSM joint angles until the measured joint torques reached the system limits. Figure~\ref{fig_rom_test} compares the model predictions with experimental results across three trials, with the torque limits of each joint indicated by red dashed lines. On average, the maximum and minimum measured outer yaw angles were $34.55^\circ$ and $-26.56^\circ$, respectively, deviating from the model predictions by only $2.71^\circ$ and $2.19^\circ$. The maximum and minimum measured outer pitch angles were $31.56^\circ$ and $-27.51^\circ$, respectively, deviating from the model predictions by only $5.5^\circ$ and $1.45^\circ$. We note that $\theta_{\mathrm{offset}}$ is nonzero only in the $x_0$--$z_0$ plane due to the weight asymmetry of the actuation assembly, resulting in a small, nonzero positive outer yaw torque of $\tau=0.3059$~Nm at zero joint angle. In contrast, $\theta_{\mathrm{offset}}=0^\circ$ in the $y_0$--$z_0$ plane, resulting in zero outer pitch torque at zero joint angle.

\section{Surgical Teleoperation Use Cases}\label{sec_teleop_demo}

In this section, we present two surgical use cases of the proposed actuation platform where teleoperation is the common control paradigm. For each use case, we first describe the continuum robot design and mechanical implementation, followed by the teleoperation strategy for user control through the input device, referred to as the Master Console Manipulator (MTM). We then discuss observations from each use case and potential directions for improving teleoperation performance. Related videos are provided in Supplementary Video 1, and the mechanical components used in the demonstrations are provided in Supplementary File 1.

\subsection{Clinical Motivation}

Robotic laparoscopic systems constrain surgical tools to pivot about a remote center of motion (RCM) at the trocar. With conventional straight, wristed instruments and cameras, this constraint produces a fulcrum effect that limits access to and reach within constrained workspaces~\cite{robotic_surgery_overview}. A continuum robot's ability to curve around obstacles can extend reach and improve access to regions that are difficult to reach with these straight tools. In addition, small-diameter continuum robots, such as CTRs with diameters of 1-2~mm, can enable manipulation in confined workspaces where instrument size limits access and dexterity. For comparison, the smallest commercially available wristed instruments have a diameter of 5~mm~\cite{Maguire2025-miniaturized-wrist}.

To demonstrate these capabilities, we present two surgical teleoperation use cases using the proposed actuation platform: (1) an 8~mm diameter TDCR camera for navigating a constrained environment and (2) a 2.53~mm diameter CTR-based forceps for miniaturized manipulation in a confined workspace alongside a 8~mm wristed forceps.

\begin{figure}[t!]
\centerline{\includegraphics[width=1.0\linewidth]{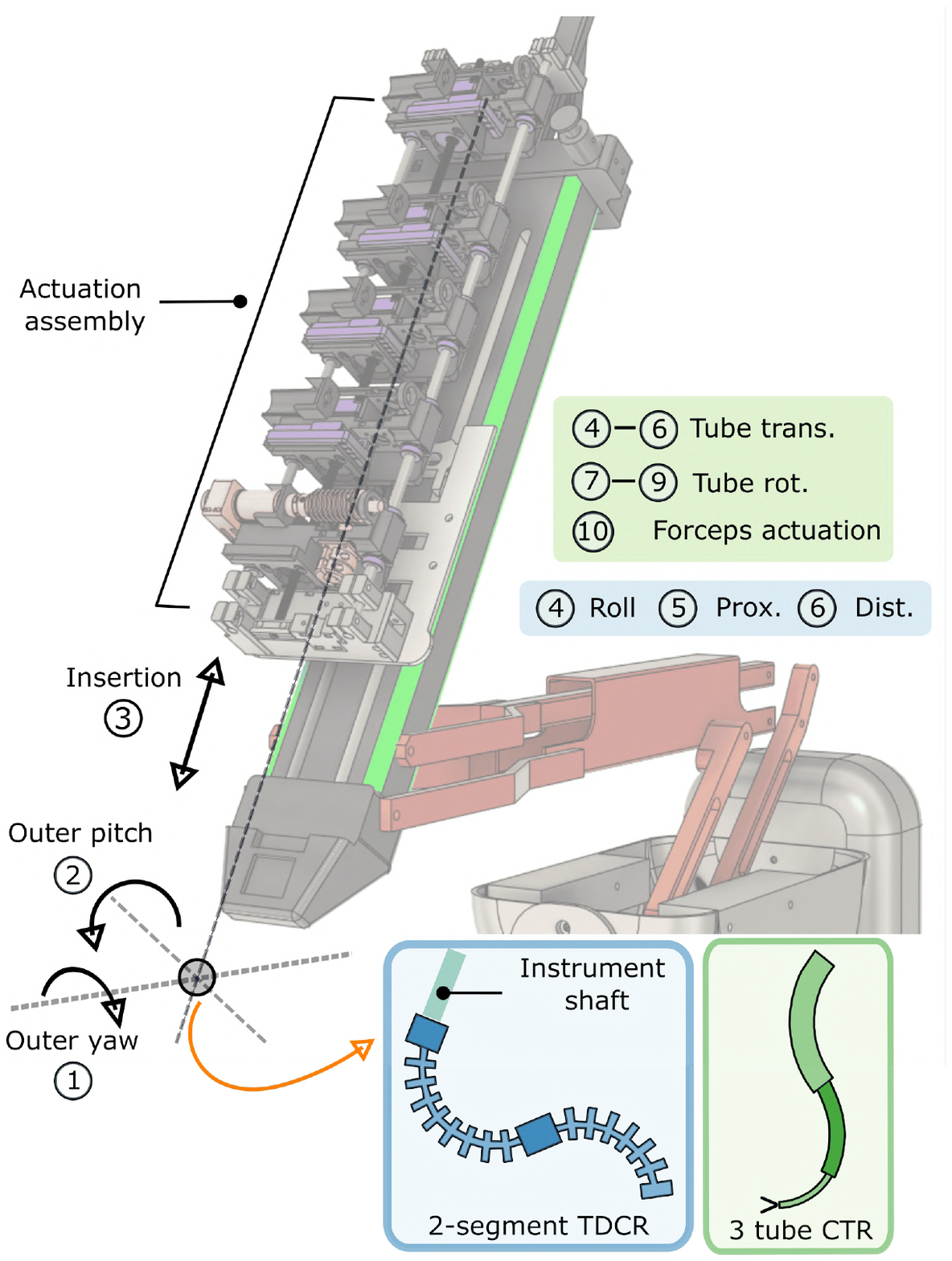}}
\vspace{-8pt}
\caption{Kinematic diagram illustrating the DOFs of two CR-dVRK architectures: a 6 DOF TDCR camera and a 9 DOF miniaturized CTR forceps instrument. Both configurations share the first three PSM joints, while the remaining actuated joints are determined by the selected continuum tool, enabled by the modular design of the proposed platform.}
\label{fig_dof_demo} 
\end{figure}

\subsection{Extending Camera Control Capabilities with a TDCR-based Camera}\label{sec_camera_control}

\subsubsection{Robot Architecture} \label{sec_TDCR_robot_architecture}
The TDCR camera had $6$ DOFs: the first $3$ were provided by the PSM, while the remaining $3$ were provided by a two-segment planar TDCR with system roll, as shown in Fig.~\ref{fig_dof_demo}.
The $3$-DoF antagonistic TDCR used $5$ actuation modules: $1$ rotation-only module (light red in Fig.~\ref{fig_module_design}) for system roll and $4$ translation-only modules for tendon actuation. We actuated each continuum segment antagonistically by equally pulling and releasing the opposing tendons. We 3D-printed the TDCR using thermoplastic polyurethane (TPU) following the fabrication process in~\cite{mangan2025serially}. The proximal segment connected to an instrument shaft that housed a 3D-printed tendon router, which separated the four tendons and prevented maypoling during system roll. We also routed the tendons individually through holes in the actuation modules to prevent entanglement and maintain a clear working channel. Finally, we integrated a $2.8$~mm-diameter camera (AMC0115USB-D2.8, Dothecamera) through the central working channel and secured it at the TDCR tip.

\subsubsection{TDCR Teleoperation Scheme} \label{sec_TDCR_teleop_mapping}
We implemented two control modes for the TDCR camera: gross camera motion and fine camera motion. In the gross motion mode, users command camera motion by actuating the PSM DOFs and system roll in joint space. In the fine motion mode, users perform fine camera adjustments by directly articulating the proximal and distal segments of the TDCR. To support both control modalities, we adapted the MTM-to-camera mapping from the dVRK teleoperation software stack~\cite{Kazanzides2014-dVRK-OG} for gross positioning and mapped the wrist pitch angles of the left MTM and the right MTM to the total bend angles of the proximal and distal segments of the TDCR, respectively. To maintain a user control experience consistent with the dVRK, we implemented a state machine that enables transitions between the two control modes, as well as hold and clutch modes.

\subsubsection{Discussion} \label{sec_TDCR_demo_discussion}

\begin{figure*}[t!]
\centering
\includegraphics[width=\linewidth]{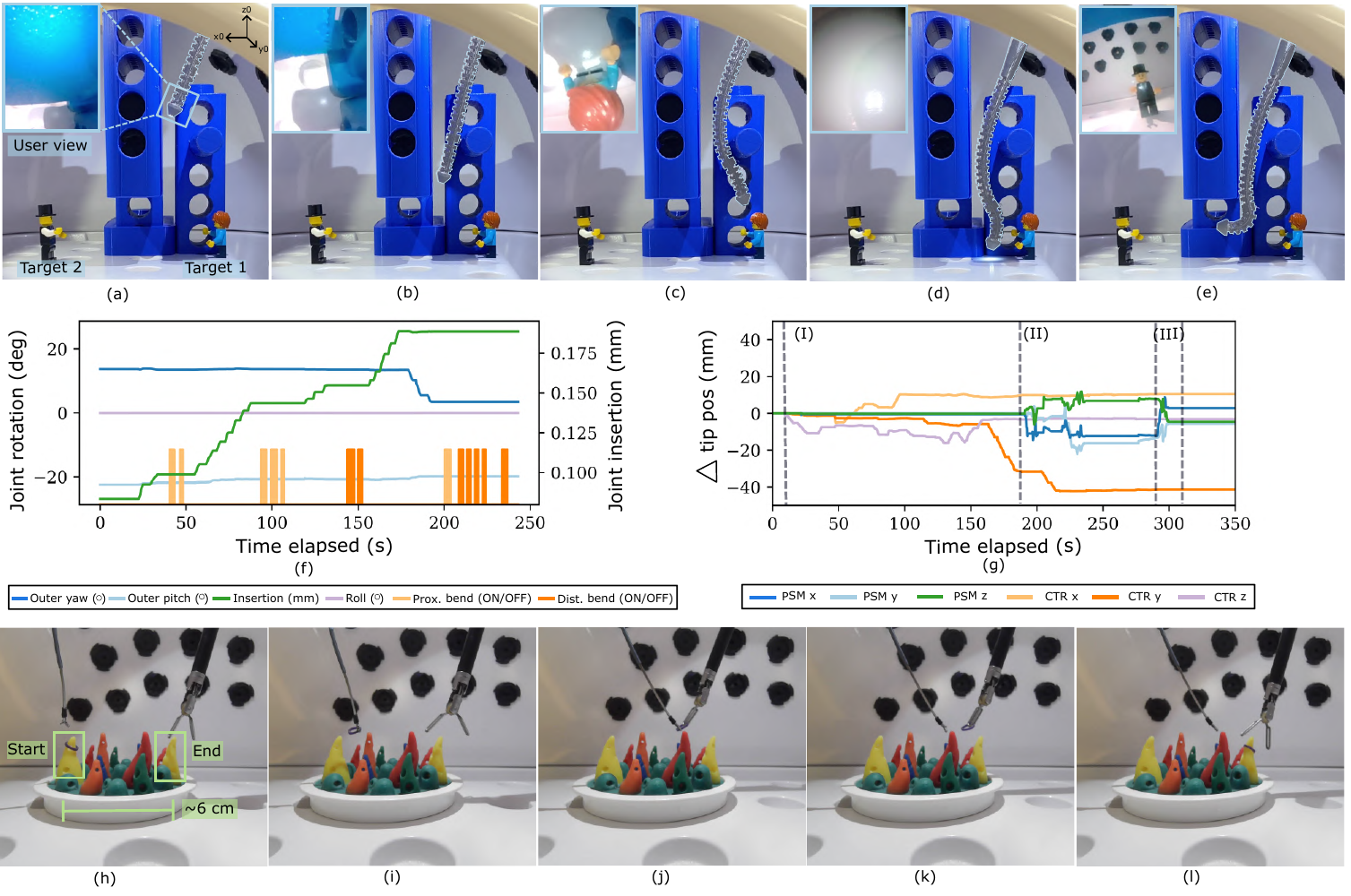}
\vspace{-12pt}
\caption{(\textit{\textbf{Top}}) (a–e) Video snapshots of TDCR-based camera navigation around obstacles, represented by a 3D-printed blue fixture, to reach two distinct targets. Targets 1 and 2 are shown in the insets of (c) and (e), respectively, and require substantial proximal and distal bending of the TDCR for visualization. The TDCR body is false-colored to improve visual contrast. (\textit{\textbf{Middle}}) (f) Time histories of PSM and TDCR joint motions. Periods of TDCR actuation are indicated by light and dark orange ON/OFF signals for the proximal and distal segments, respectively. (\textit{\textbf{Middle}}) (g) Time histories of the change in tip positions of the CTR and PSM. Distinct phases of CTR manipulation, handoff, and PSM manipulation are labeled with I, II, III, respectively. (\textit{\textbf{Bottom}}) (h–l) Video snapshots of a peg-transfer task performed by the CTR-based forceps in coordination with an 8~mm wristedforceps.}
\label{fig_teleop_demo_combined}
\end{figure*}

As shown in Fig.~\ref{fig_teleop_demo_combined} (a)-(e), the targets are positioned such that actuating both bending segments of the TDCR is required for proper visualization within the obstacle environment. Fig.~\ref{fig_teleop_demo_combined} (f) presents the time histories of individual joint values during the task. To visualize the first target, the proximal segment must bend around the rear peg during scope insertion (see Fig.~\ref{fig_teleop_demo_combined} (b)–(c)). Fig.~\ref{fig_teleop_demo_combined} (f) further shows that, from $0s$ to $120s$, insertion increases continuously with intermittent activation of the TDCR proximal bending segment, indicated by a light orange ON/OFF signal. We note that this maneuver is not feasible with a straight camera, as rotating the entire scope around the RCM about the $y0$ axis toward the first target would cause the scope body to collide with the rear peg, thereby preventing further camera motion.

Visualizing the second target requires a coordinated combination of TDCR and PSM joint motions (see Fig.~\ref{fig_teleop_demo_combined} (d)–(e)). Specifically, the proximal and distal segments must bend in tandem to form an S-shaped configuration that passes beneath the wall ahead; this is evident from $120s$ to $250s$ in Fig.~\ref{fig_teleop_demo_combined}(f), shown by both light and dark orange ON/OFF signals. Because the targets are positioned out of plane along the $y0$, a rotation of the entire PSM about the $x0$ axis must be achieved to visualize the second target, as shown by the outer yaw joint value between $150s$ to $200s$. 


\subsection{Bimanual Teleoperation with a CTR-based forceps}

\subsubsection{Robot Architecture}
The CTR-based forceps had $10$ DOFs: $3$ from the PSM, $6$ from a three-tube CTR, and $1$ from forceps actuation. 
The $7$-DOF CTR-based forceps required three actuation modules, each capable of linear translation and axial rotation. We retrofitted the most proximal module with an additional instrument mechanism for forceps actuation (see section~\ref{sect_instrument_actuation}). 
A shaft collar (Fig.~\ref{fig_module_design}) secured each tube to its corresponding module, and 
Table~\ref{table_CTR_physical_parameters} summarizes the parameters for the CTR tubes.
We routed a modified forceps (EndoJaw, Olympus) through the working channel (orange in Fig.~\ref{fig_module_design}) to enable distal grasping.

\begin{table}
    \centering
    \caption{CTR Tube Parameters}  
    \begin{tabular}{cccc}
        Variable & Tube 1 &  Tube 2 &  Tube 3\\
        \hline
         $ID (mm)$& 0.966  & 1.460  & 2.21\\
         $OD (mm)$ & 1.270  &  2.032 & 2.53\\
         $L_s (mm)$& 230 &  135& 60\\
         $L_c (mm)$ &  40& 55 & 70\\
         $\kappa (mm^{-1})$& 0.01  & 0.01 & 0.004\\
         \hline
    \end{tabular}
    \label{table_CTR_physical_parameters}
\end{table}

\subsubsection{CTR Teleoperation Scheme} \label{sec_CTR_teleop_mapping}
We implemented Cartesian tip position control for the CTR by mapping the MTM’s relative gripper translation to the CTR’s relative translation, enabling fine control of the tip motion. In addition, we enabled forcep opening and closing by directly mapping this action to the MTM gripper’s open and close motion.

To achieve real-time tip position control, we approximated the torsionally compliant kinematics of the CTR using a Fourier series representation~\cite{Dupont2009-design-control-ctr}. Specifically, we employed a Fourier series of order $2$ for each of the 6 joints, resulting in $15625$ coefficients. To limit computational memory requirements, we used $75,000$ tip position samples generated from a Cosserat rod–based CTR model~\cite{Rucker-CTR-modeling} to train the coefficients. The resulting approximation achieved a mean $2$-norm position error of $0.246\text{mm}$ when evaluated on a separate test set of 
$150,000$ samples. To drive the CTR tip to its desired position, we solved the inverse kinematics by adapting the constrained quadratic program from~\cite{Xu2024-shape-control}: 

{\small
\begin{equation}\label{tipleastsquares}
\begin{aligned}
        \min_{\Delta q} \quad & \frac{1}{\lVert p_0 \rVert^2}\frac{1}{2} \lVert J\Delta q - \Delta p \rVert^2\\
        \text{s.t.} \quad 
        & (C_1) \quad l_1\le l_2 + \epsilon \le l_3 + \epsilon ,\\
        & (C_2) \quad \Delta \underline{q} \leq \Delta q \leq \Delta \bar{q}. 
\end{aligned}
\end{equation}
}

At each teleoperation step, the program computes the optimal joint increment $\Delta q$ that minimizes the difference between the desired tip position displacement $\Delta p$ and linearized kinematic mapping described by the tip Jacobian $J$, normalized by $\lVert p_0 \rVert$. 
Constraint $C_1$ enforces a minimum separation of $\epsilon$ between tube deployments $l_{1,2,3}$, while constraint $C_2$ limits tube translation and rotation velocities to feasible values at every control cycle, thereby mitigating the effects of motor lag~\cite{Xu2024-shape-control}. 
In addition, one can toggle between clutch, hold, and CTR task space control mode via a state-machine.

\subsubsection{Discussion} \label{sec_CTR_demo_discussion}
As shown in Fig.~\ref{fig_teleop_demo_combined}~(h–l), the $2.53$~mm CTR-based forceps successfully grasped a rubber band from one peg, transferred it to a $8$~mm wristed forceps, and placed it onto another peg approximately $6$~cm away, demonstrating the feasibility of teleoperated pick-and-place tasks at a miniaturized scale. Figure~\ref{fig_teleop_demo_combined}(g) shows the corresponding time histories of the tip positions of the CTR and wristed forceps, with the three previously described periods labeled (I), (II), and (III), respectively.


The proposed CTR-based forceps are well-suited for procedures with extremely small and constrained operating workspaces. For example, in pediatric procedures, the available workspace can be as small as 3-4~cm$^3$~\cite{Thakre2008-uu,NOWAK1994845}, where accommodating even the smallest possible 5~mm instrument can be challenging, let alone multiple instruments. In such settings, instrument miniaturization is critical for enabling maneuverability while avoiding collisions, particularly in multi-instrument surgical scenes.


\subsection{Extension to Other Continuum Robot Designs and Applications}
\begin{figure}[t!]
\centering
\includegraphics[width=\columnwidth]{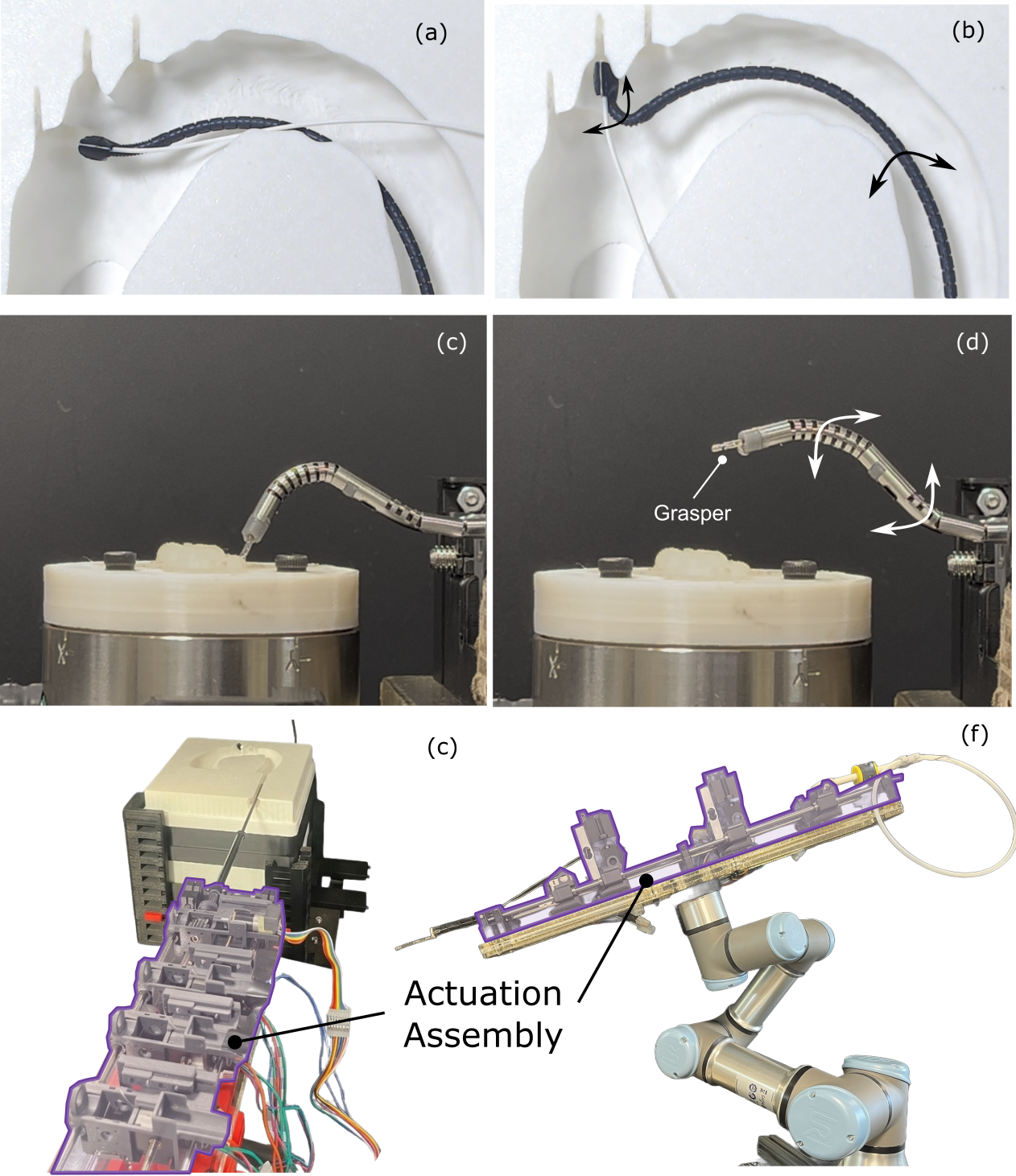}
\vspace{-12pt}
\caption{The proposed actuation platform in different applications and mounting configurations. \textit{\textbf{Top}} (a--b): autonomous navigation of a notched TDCR around the aortic arch~\cite{mangan2026contact}; \textit{\textbf{Middle}} (c--d): continuum robotic grasper for submucosal resection; \textit{\textbf{Bottom}} (e--f): mounting configurations on a linear stage and a UR robotic arm.}
\label{fig_actuation_examples}
\end{figure}

The proposed actuation platform can be even further extended beyond the continuum robot designs and surgical applications demonstrated in this work. To date, the platform has also been used as a standalone system mounted on a linear stage to actuate various tendon-driven continuum robot designs for surgical applications (see Fig.~\ref{fig_actuation_examples}). These include a $2$-DOF continuum robot for navigation through the aortic arch~\cite{mangan2026contact} (Fig.~\ref{fig_actuation_examples} (a)-(b)) and a $3$-DOF continuum robotic grasper for surgical resection tasks (Fig.~\ref{fig_actuation_examples} (c)-(d)). In addition, the proposed actuation platform can also be integrated with commercial serial manipulators, such as the Universal Robots (UR) platform, providing an alternative host robotic arm (Fig.~\ref{fig_actuation_examples} (f)). These examples further demonstrate the potential applicability of the proposed platform across different continuum robot architectures, DOFs, and task applications.

\section{Conclusion}
In this work, we presented a unified, modular actuation platform that decouples continuum robot actuation from robot design through integrated mechanical, electrical, and software components. The platform supports multiple continuum robot classes with arbitrary DOFs and hybrid continuum–serial architectures. Characterization experiments demonstrated sub-millimeter translational accuracy and rotational accuracy below $5^\circ$. We validated the platform on a hybrid continuum–serial robot system, demonstrating two representative teleoperated surgical use cases. Future work will explore integrated sensing, improved teleoperation mappings, and redundancy resolution to enable advanced behaviors such as follow-the-leader motion.


\section*{Acknowledgments}
The authors would like to thank Noah Jones for his contributions to the electrical integration and initial motor motion control implementation; Sara Sariowan for her contributions to the dVRK attachment design; and Jui-Te Lin for his initial guidance on the instrument actuation design.

\bibliographystyle{IEEEtran}
\bibliography{biblio}

\vskip -2\baselineskip plus -1fil

\begin{IEEEbiography}[{\includegraphics[width=1in,height=1.25in,keepaspectratio]{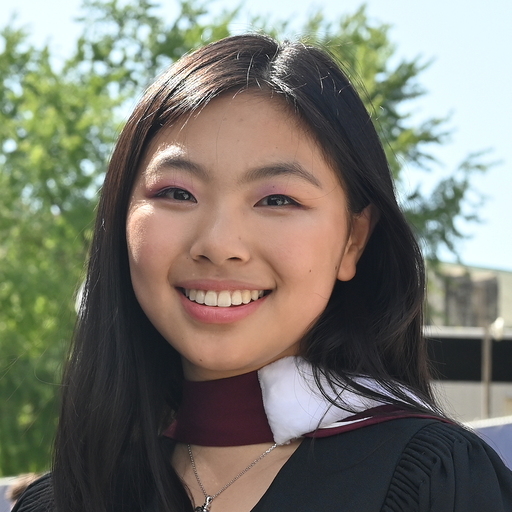}}]{Yunti Xu} (Graduate Student Member, IEEE) received the B.Sc. degree in Engineering Science from University of Toronto, ON, Canada in 2022. She is currently pursuing the Ph.D. degree in the Department of Mechanical and Aerospace Engineering, University of California, San Diego, CA, USA, with a focus on system design and control of continuum robot systems for surgical applications. Her research interests include continuum robot control and teleoperated surgical systems.
\end{IEEEbiography}

\vskip -2\baselineskip plus -1fil

\begin{IEEEbiography}[{\includegraphics[width=1in,height=1.25in,keepaspectratio]{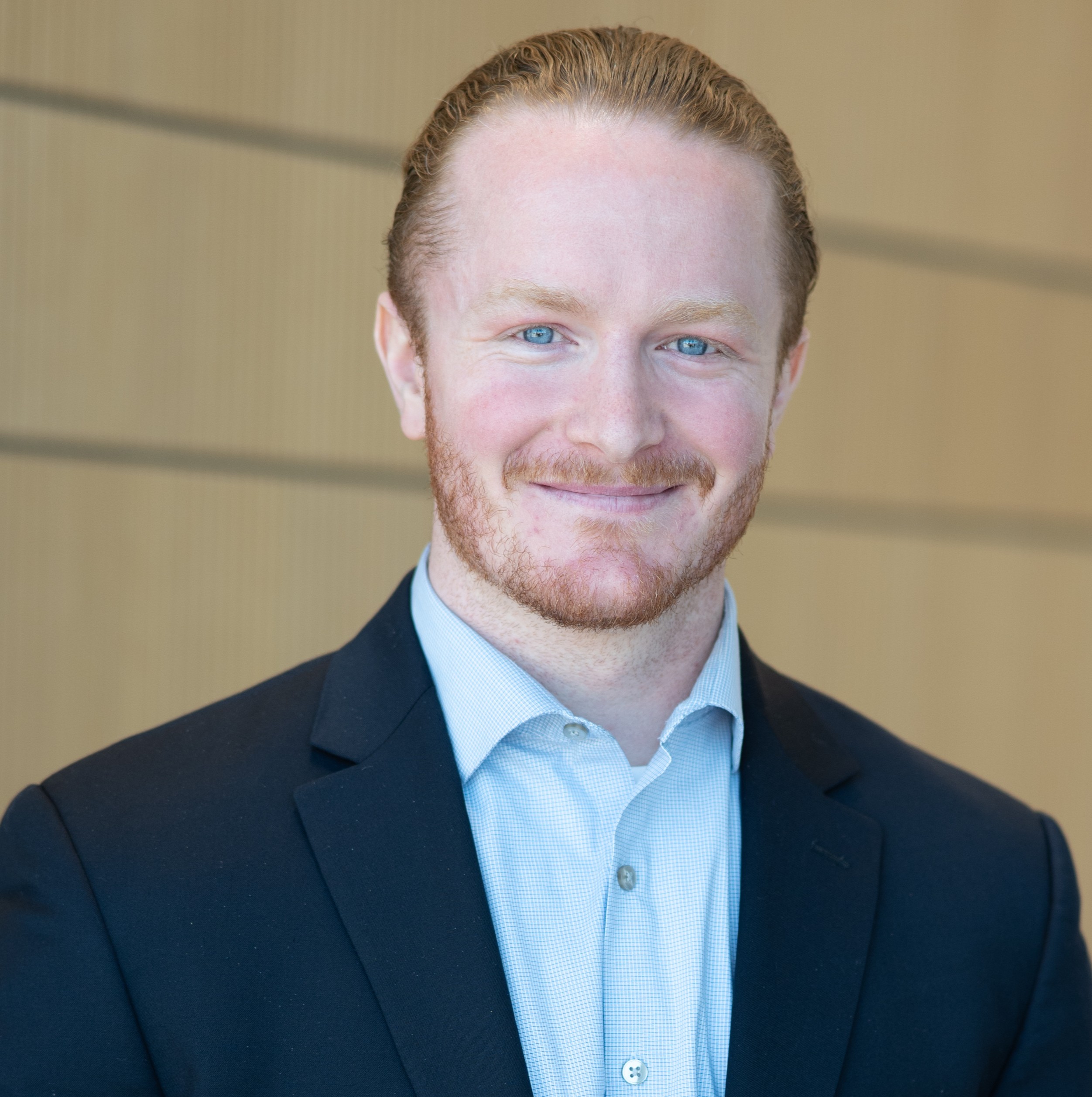}}]{Aedan Mangan} (Graduate Student Member, IEEE) received a Bachelor’s degree in Mechanical Engineering from the University of Dayton, Dayton, OH, USA, and a Master’s degree in Bioengineering from Rice University, Houston, TX, USA. He worked full-time at Johnson \& Johnson, where he contributed to the development of the OTTAVA Robotic Surgical System, and is currently pursuing a Ph.D. degree in Mechanical and Aerospace Engineering at the University of California San Diego, La Jolla, CA, USA. He is also a Visiting Graduate Student with the Department of Computer Science at the University of Toronto, Toronto, ON, Canada. His research interests include design, planning, control, and estimation for continuum and medical robots.
\end{IEEEbiography}

\vskip -2\baselineskip plus -1fil

\begin{IEEEbiography}[{\includegraphics[width=1in,height=1.25in,keepaspectratio]{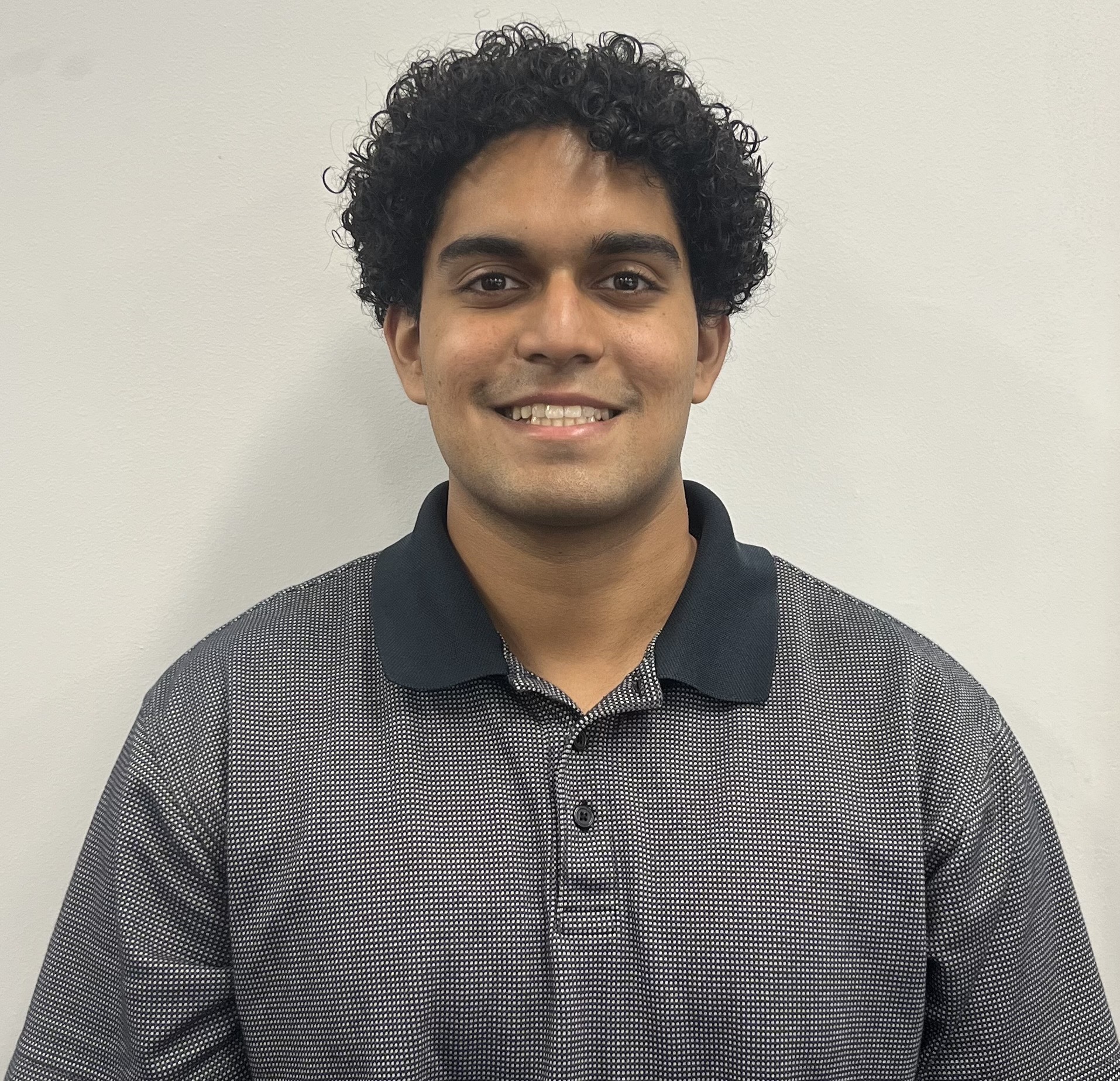}}]{Miheer Potdar} (Student Member, IEEE) received the B.S. degree in Mechanical Engineering from University of California, San Diego, CA, USA in 2026. His research interests include robot design, mechatronics, and medical robotic systems. He is currently at Tesla, working on cell manufacturing for the 4680 program.
\end{IEEEbiography}

\vskip -2\baselineskip plus -1fil

\begin{IEEEbiography}[{\includegraphics[width=1in,height=1.25in,keepaspectratio]{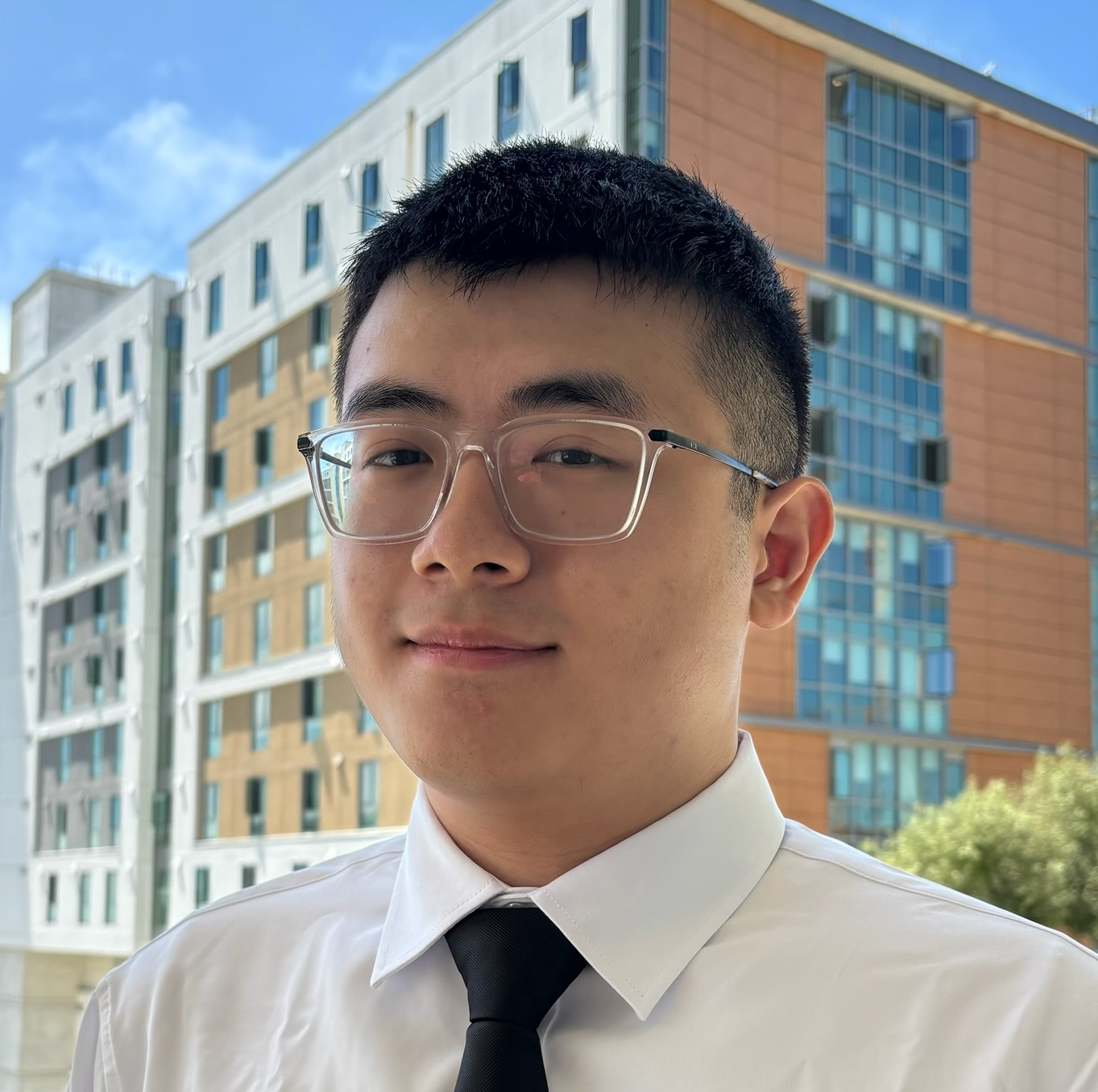}}]{Alex Lu} (Student Member, IEEE) Alex Lu received the B.S. degree in bioengineering and the M.Eng. degree in medical device engineering from the University of California, San Diego, CA, USA. He has academic experience in the mechanical design, iterative prototyping, and system validation of robotic instruments. His research interests include continuum robots, modular actuation systems, and smart medical devices.
\end{IEEEbiography}

\vskip -2\baselineskip plus -1fil

\begin{IEEEbiography}
[{\includegraphics[width=1in,height=1.25in,clip,keepaspectratio]{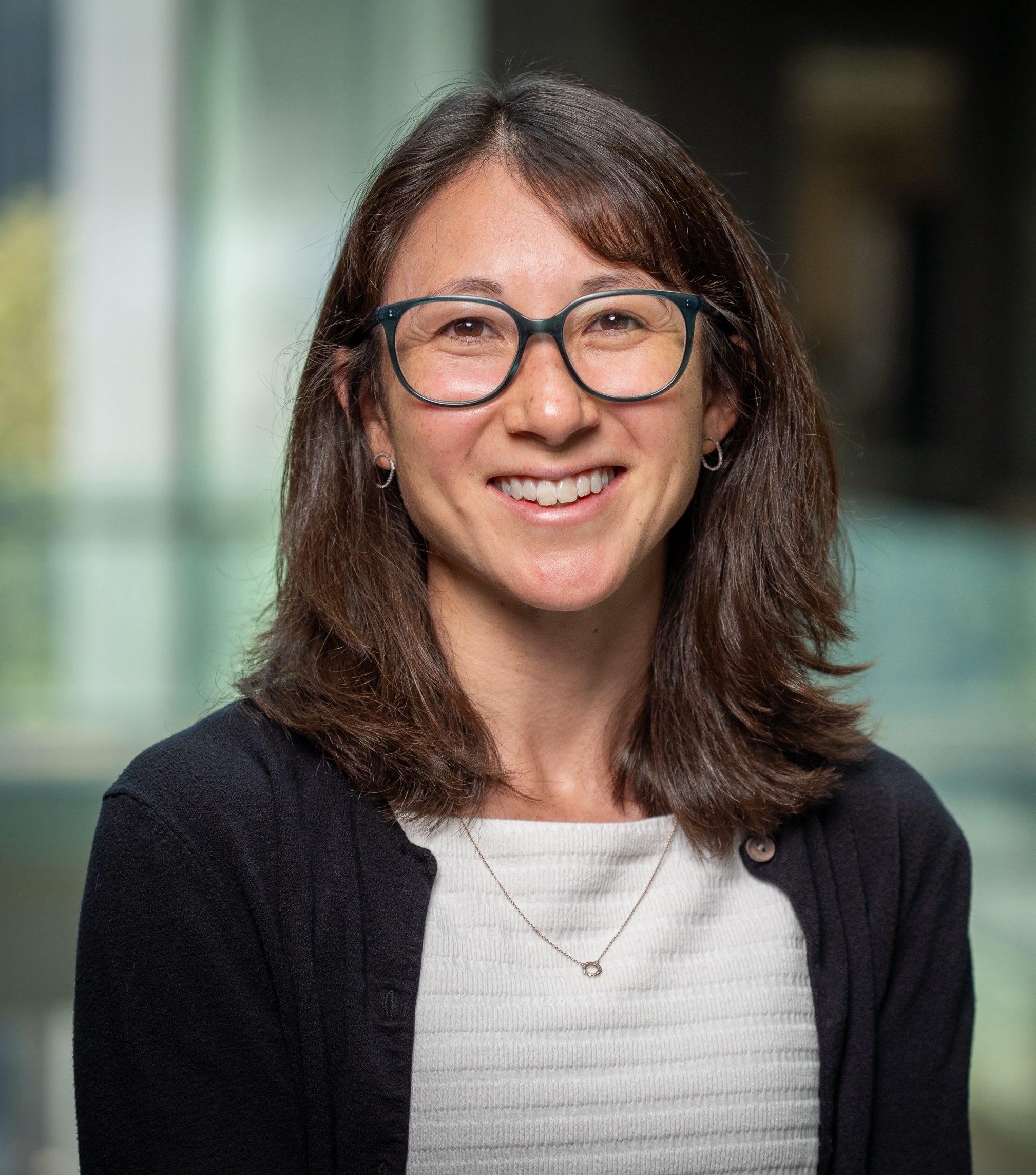}}]{Tania K. Morimoto} (Senior Member, IEEE) received the B.S. degree from Massachusetts Institute of Technology, Cambridge, MA, in 2012 and the M.S. and Ph.D. degrees from Stanford University, Stanford, CA, in 2015 and 2017, respectively, all in mechanical engineering. She is currently an Associate Professor in the Department of Mechanical and Aerospace Engineering at University of California, San Diego. Her research interests include robotics, haptics, and engineering education. Dr. Morimoto was the recipient of the NSF CAREER Award (2022) and the Beckman Young Investigator Award (2022).
\end{IEEEbiography}

\vfill

\end{document}